\documentclass{article}
\usepackage[preprint]{spconf}
\usepackage{hyperref}       
\usepackage{url}            
\usepackage{booktabs}       
\usepackage{amsfonts}       
\usepackage{amsmath}
\usepackage{amssymb}
\usepackage{nicefrac}       
\usepackage{microtype}      
\usepackage{cite}
\usepackage{xcolor}
\usepackage{graphicx,subfigure}
\graphicspath{ {./figures/} }

\title{EMPIRICAL ANALYSIS OF OVERFITTING AND MODE DROP IN GAN TRAINING}
\name{Yasin Yazici$^{\star}$$^{\dagger}$ \qquad Chuan-Sheng Foo$^{\dagger}$ \qquad Stefan Winkler$^{\ddagger}$ \qquad Kim-Hui Yap$^{\star}$ \qquad Vijay Chandrasekhar$^{\dagger}$\sthanks{Contributed in 2019.}}

\address{$^{\star}$ School of Electrical and Electronic Engineering, Nanyang Technological University, Singapore \\
$^{\dagger}$ Institute for Infocomm Research, A*STAR, Singapore \\
$^{\ddagger}$ School of Computing, National University of Singapore
}

\begin{document}
%
\maketitle
\copyrightnotice{\copyright\ IEEE 2020}
\toappear{To appear in {\it Proc.\ ICIP2020, October 25-28, 2020, Abu Dhabi, United Arab Emirates}}

\begin{abstract}
We examine two key questions in GAN training, namely overfitting and mode drop, from an empirical perspective. We show that when stochasticity is removed from the training procedure, GANs can overfit and exhibit almost no mode drop. Our results shed light on important characteristics of the GAN training procedure. They also provide evidence against prevailing intuitions that GANs do not memorize the training set, and that mode dropping is mainly due to properties of the GAN objective rather than how it is optimized during training.
\end{abstract}
\section{Introduction}
\label{sec:intro}
We perform empirical analyses to address key questions relating to overfitting, generalization and mode dropping in the training of generative adversarial networks (GAN). We hypothesize that these phenomena are related to stochasticity in GAN training and provide experimental evidence to support this intuition. We show that specific GAN architectures and optimization methods can overfit to the training set. We define overfitting as the case when the generator produces images that are nearly indistinguishable from the (training) data samples and ``covers'' all of them. By definition such a generator does not exhibit mode drop as it has complete recall on the training data distribution. Our experiments provide an answer to whether the GAN objective and its variants are sufficient to match the support of the training data distribution as questioned by \cite{arora2018do}. It leads to new insights into how GANs generalize and why mode drop occurs. 

While our observation that GANs overfit when stochasticity is removed may appear obvious by analogy to the behavior of models trained with maximum log-likelihood objectives, GANs are trained by optimizing a saddle-point objective and exhibit very different training dynamics. In particular, variants of gradient descent on the saddle-point objective
may not converge without regularization even in simple cases \cite{MeschederICML2018}.

In summary, our contribution is to empirically show that in GAN training (a) the generator tends to overfit to a large extent as stochasticity decreases, and similarly (b) the generator shows limited mode drop behavior, which reveals the relationship between mode drop and stochasticity.

\section{Related Work}

\textbf{Overfitting:} Do GANs overfit or memorize the data in the training set? This question is answered negatively in many papers, by searching for nearest-neighbors of generated images in training dataset \cite{gan,brock2018large,karras2018progressive}. Using a different approach, \cite{DBLP:journals/corr/abs-1901-03396,DBLP:conf/iclr/MetzPPS17} analyzed overfitting in GANs and other generative models through searching in the code space of the generator. While the latter approaches retrieve generated samples closer to ones in the training set, there are significant differences between the images, especially in the fine details. \cite{DBLP:journals/corr/abs-1902-03468} suggests that memorization may not be happening as the generator does not directly learn from the training samples but from the feedback of the discriminator. These works suggest that GANs do not overfit to the training set.\footnote{~\cite{brock2018large} claims that the discriminator overfits by showing the discrepancy between the logits of training set and validation set. In this paper, we are looking at overfitting/memorization from the generator side not the discriminator.}


\textbf{Mode drop (or collapse):} This is a notorious behavior of GANs; instead of covering the full support of the data distribution, GANs tend to cover only parts of it. Some link this behavior to the underlying divergence of the GAN objective (non-saturating version) which is related to the reverse-KL divergence \cite{pmlr-v70-arora17a, DBLP:conf/iclr/ArjovskyB17}. Theoretical studies suggest that the low capacity of the discriminator is related to mode dropping \cite{pmlr-v70-arora17a}. \cite{arora2018do} analyzed this empirically using a birthday paradox test and concluded that GANs are prone to leaving modes out. \cite{NIPS2018_7495} related mode drop to the mismatch between the multimodality of data distribution and the unimodality of the prior distribution of the generator. Motivated by disconnected manifolds, \cite{NIPS2018_7964} also blames the multimodality of the data distribution and suggests a mixture distribution of multiple generators. \cite{DBLP:journals/corr/abs-1807-04015} interprets mode drop with catastrophic forgetting as a result of continuously changing generative distribution. By contrast, we show that mode drop is avoidable to a large extent by simply removing stochasticity from the training, while keeping the prior distribution and objective function the same. 


\cite{2019arXiv190408598C} suggested methods to reduce variance in gradient calculations without using larger mini-batches. In essence, we also reduce stochasticity by means of larger batch sizes. However the aim of our paper is different and more generally shows how overfitting and mode dropping are linked to stochasticity in training. \cite{gan_memorization} provides a theoretical analysis of GAN memorization, which is defined there as the case when the generator distribution matches the empirical data distribution over all samples from the prior, whereas our definition requires a match only on limited samples from the prior; \cite{gan_memorization} also does not aim to answer questions about mode dropping.

\section{Method}


GAN is a two player zero-sum game between a discriminator and a generator:
\begin{equation*}
    \min_{G} \max_{D} E_{\boldsymbol x \sim p_{data}(\boldsymbol x)}[\log D(\boldsymbol x)] +  E_{\boldsymbol z \sim p_{z}(\boldsymbol z)}[\log(1-D(G(\boldsymbol z)))]
    \label{minimax}
\end{equation*}

It utilizes a discriminator to assess a peudo-divergence between the true data distribution, $p_{data}(\boldsymbol x)$, and the generator's distribution, $p_{g}(\boldsymbol x)$. The discriminator maximizes the divergence, while the generator minimizes it. In this way, the generator learns to mimic the data distribution implicitly. 


In practice, the objective function is approximated with empirical averages:
\begin{equation*}
    \min_{G} \max_{D}\frac{1}{n} \sum_{i=1}^{n} \log D(\boldsymbol x_{i}) +  \frac{1}{k} \sum_{i=1}^{k} \log(1-D(G(\boldsymbol z_i)))
    \label{minimax_emp}
\end{equation*}
where $\boldsymbol x_i$ denotes the $i^\mathrm{th}$ sample from a fixed set $ \boldsymbol X = \left\{ \boldsymbol x_1,..., \boldsymbol x_n \right \}$ sampled from  $p_{data}(\boldsymbol x)$, and $\boldsymbol z_i$ is the $i^\mathrm{th}$ sample from a fixed set $ \boldsymbol Z = \left\{ \boldsymbol z_1,...,\boldsymbol z_k \right \}$ sampled from input noise $p_{z}(\boldsymbol z)$. In practice $k>>n$, as $p_{z}(\boldsymbol z)$ is readily available, while data samples are limited.


During training, gradients of the objective are further approximated with stochastic gradients using a mini-batch of size $m$, leading to the following parameter updates:
\begin{align*}
    \phi^{(t+1)} &= \phi^{(t)} + \frac{\alpha}{m} \sum_{i=1}^{m} \frac{\partial (\log D(\boldsymbol x_{i}) + \log(1-D(G(\boldsymbol z_i)))) }{\partial \phi^{(t)}} \\
    \theta^{(t+1)} &= \theta^{(t)} - \frac{\alpha}{m} \sum_{i=1}^{m} \frac{\partial (\log(1-D(G(\boldsymbol z_i))))}{\partial \theta^{(t)}}
\end{align*}
where $m$ samples are drawn from $ \boldsymbol X$ with replacement and from $ \boldsymbol Z$ without replacement,\footnote{As new samples are drawn from $p_{z}(\boldsymbol z)$ at each iteration, and it is unlikely to draw the exact same sample again.} $\phi^{(t)}$ and $\theta^{(t)}$ are the parameters of $D$ and $G$ at the $t^\mathrm{th}$ iteration respectively. 

There are two sources of stochasticity in GAN training: (i) random samples from the prior and (ii) stochastic gradient updates.\footnote{We did not use stochastic components in the network (like batch-norm).} The former is considered as a necessary step to capture support of the prior distribution, while the latter is used to reduce computation time by approximating the true gradient. To remove both sources of stochasticity, we select $k=n$ and $m=n$. Under these settings, $G$ generates exactly $n$ samples during training, and $D$ only distinguishes them from $n$ samples of the empirical data distribution. As $G$ only generates a finite number of samples, this enables us to analyze pairwise distances between generated samples and data samples (or vice versa). 

After analyzing the deterministic setting, we re-introduce varying amounts of stochasticity to investigate its effect on overfitting, generalization, and mode drop. We re-introduce stochastic gradients by reducing the mini-batch size $m$ to $\frac{n}{2^l}, l={2,4,6,8}$. We also re-introduce noise into the input (code) space of the generator. We did not consider sampling more from the prior as it modifies the ratio of
real/fake samples. Each time, a single type of stochasticity is considered to study its effect in isolation.

Our intuition is that stochastic optimization leads the generator to leave out some modes and reduces the fidelity of image generation. In the initial training phase, stochasticity from the gradients due to sampling error is smaller than the estimated difference between the true and learnt distributions which enables GAN training to progress. However as training progresses, the generator manifold gets closer to the data manifold and stochasticity dominates the training, forcing the discriminator to find superficial explanations for the difference between the distributions. This prevents further progress in learning the data distribution and leads to non-convergence over the parameters. Indeed, the performance of features extracted from the GAN's discriminator deteriorates for a downstream classification task in later stages of training \cite{DBLP:journals/corr/abs-1811-11212}. Also, large batch size training shows significant improvements in image generation \cite{brock2018large}.


\section{Experiments}\label{experiments}

We perform experiments on the SVHN~\cite{svhn}, CIFAR-10~\cite{cifar10} and FFHQ~\cite{DBLP:journals/corr/abs-1812-04948} datasets commonly used for evaluating GANs. FFHQ images are scaled to $32 \times 32$ to make the analysis possible with large batch sizes and sufficiently large data sizes. We use 12,800 images from each dataset for training. In all our experiments, we use the non-saturating GAN loss \cite{gan} with alternating gradient descent and a 1:1 ratio for discriminator/generator updates.

We use a DCGAN \cite{dcgan} like architecture with details in the Apendix. In the discriminator we apply spectral normalization \cite{miyato2018spectral} which improves the results considerably.  ADAM optimizer \cite{adam} with $\alpha = 0.0001$, $\beta_1 = 0$ and $\beta_2 = 0.9$ is used. Exponential moving average over generator parameters  \cite{yasin_ema,karras2018progressive} is used for evaluation and visualization. $z \in \mathbb{R}^{512}$ are sampled from $\mathcal{N}(\boldsymbol 0,\boldsymbol I)$ and fixed during training.

We include two types of randomness analysis in this paper.  We show the effects of mini-batch size as the first type of stochasticity. Then, by fixing the mini-batch size, we include noise to the fixed input samples of the generator as a second type of stochasticity. For this we replace $G(z)$ with $G(\hat{z})$ where $\hat{z} = z + \epsilon, \epsilon \sim \mathcal{N}(\boldsymbol 0,0.5\boldsymbol{I})$ during training. This corresponds to using an equally weighted mixture of $n$ Gaussians with fixed mean and co-variance. During the inference stage, when analyzing and visualizing the generator, $G(z)$ is used, which corresponds to the mean of each Gaussian. This can also be considered as most likely mode of truncation trick \cite{brock2018large}. 

To quantify overfitting and mode drop we utilize two types of metrics: (a) pixel-wise loss in data space with $\ell_1$-norm, and (b) semantic loss in a latent space with $\ell_2$-norm. The main reason to use pixel-wise loss is to show the generator can produce data samples closely. As this loss may not find semantically similar samples in certain cases~\cite{Theis2015d}, we also utilize semantic loss.  With these metrics, nearest neighbor algorithm (1-NN) is used to evaluate and visualize the distance to the first nearest neighbor. For overfitting, this analysis is done by searching the nearest neighbor of each generated samples in all data samples, while for mode drop analysis it is done in the reverse order: searching the nearest neighbor of each data sample in all generated samples. The former can be interpreted as precision and the latter as recall. 

Because search for the nearest sample is an expensive operation for large datasets, we  approximate it with $200$ samples. For example, for the overfitting analysis, the nearest neighbor of $200$ generated samples are searched in all data samples. For the semantic loss, we use \textit{InceptionV3} \cite{44903} architecture's penultimate layer activations similar to FID \cite{NIPS2017_7240}. For pixel-wise loss, images are scaled to $[0,255]$ to be able to interpret the results more easily.

The above metrics average over the samples, which does not clearly demonstrate the worst cases in that average. To include those, we also report the top 10\% and 5\% samples with highest first nearest neighbor (Table~\ref{quantitative_table}).

\section{Results}

We begin the analysis from the visual part which motivated this work. Our qualitative results are best interpreted together with the quantitative results as we can only show limited samples. For qualitative results we show both pixel-wise loss and semantic loss, while for quantitative results we only show the pixel-wise loss as it is sufficient to support our hypothesis. 

In our early experimentation we used full-batch setting without a significant difference from a batch size of 3,200 in terms of performance, thus we stick with highest mini-batch size of 3,200 to approximate the deterministic case. We train each case till convergence of our metrics. 


\subsection{Mini-Batch Size}

\begin{figure*}[h!]
    \vspace{-3mm}
    \centering
    \subfigure{\includegraphics[width=0.30\textwidth]{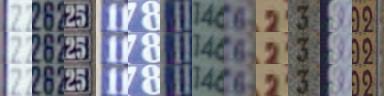}} \quad
    \subfigure{\includegraphics[width=0.30\textwidth]{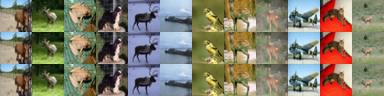}} \quad
    \subfigure{\includegraphics[width=0.30\textwidth]{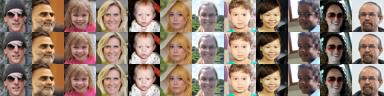}} \par
    \vspace{-3mm}
    \subfigure{\includegraphics[width=0.30\textwidth]{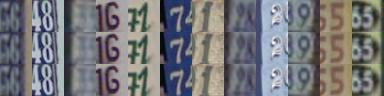}} \quad
    \subfigure{\includegraphics[width=0.30\textwidth]{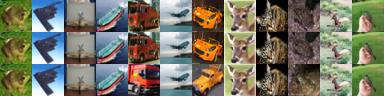}} \quad
    \subfigure{\includegraphics[width=0.30\textwidth]{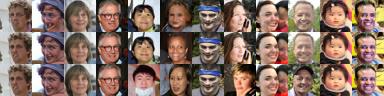}} \par
    \vspace{-3mm}
    \subfigure{\includegraphics[width=0.30\textwidth]{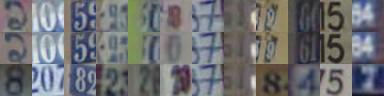}} \quad
    \subfigure{\includegraphics[width=0.30\textwidth]{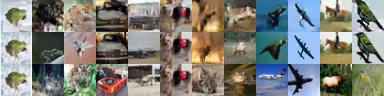}} \quad
    \subfigure{\includegraphics[width=0.30\textwidth]{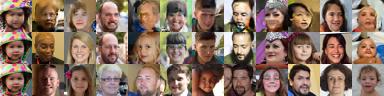}} \par
    \vspace{-3mm}
    \subfigure{\includegraphics[width=0.30\textwidth]{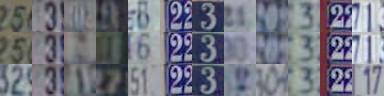}} \quad
    \subfigure{\includegraphics[width=0.30\textwidth]{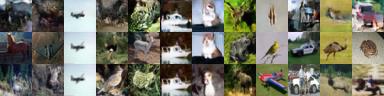}} \quad
    \subfigure{\includegraphics[width=0.30\textwidth]{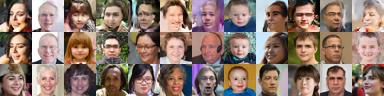}} \par
    \vspace{-3mm}
    \subfigure{\includegraphics[width=0.30\textwidth]{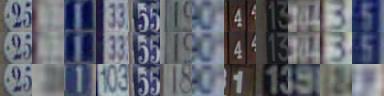}} \quad
    \subfigure{\includegraphics[width=0.30\textwidth]{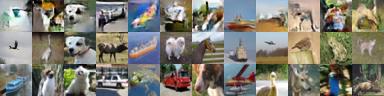}} \quad
    \subfigure{\includegraphics[width=0.30\textwidth]{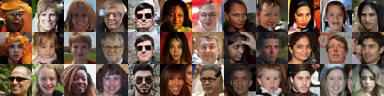}} \par
    \vspace{-3mm}
    \caption{Analysis of overfitting on SVHN (left), CIFAR10 (middle) and FFHQ (right) as a function of mini-batch size (3,200/800/200/50 from the top) and noisy latent space (bottom row). For each subfigure, the first row shows generated samples, while the second and third rows show the closest data samples in pixel space and latent space, respectively.}
    \label{fig:of}
\end{figure*}

\begin{figure*}[h!]
    \centering
    \subfigure{\includegraphics[width=0.30\textwidth]{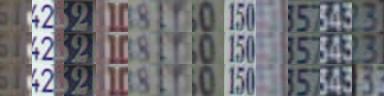}} \quad
    \subfigure{\includegraphics[width=0.30\textwidth]{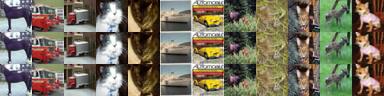}} \quad
    \subfigure{\includegraphics[width=0.30\textwidth]{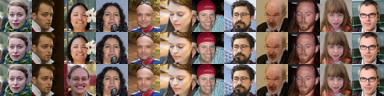}} \par
    \vspace{-3mm}
    \subfigure{\includegraphics[width=0.30\textwidth]{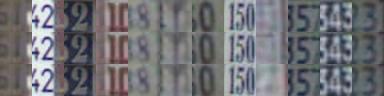}} \quad
    \subfigure{\includegraphics[width=0.30\textwidth]{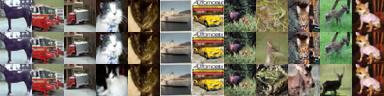}} \quad
    \subfigure{\includegraphics[width=0.30\textwidth]{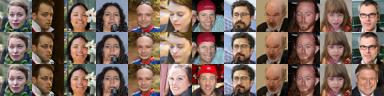}} \par
    \vspace{-3mm}
    \subfigure{\includegraphics[width=0.30\textwidth]{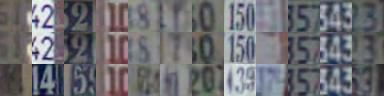}} \quad
    \subfigure{\includegraphics[width=0.30\textwidth]{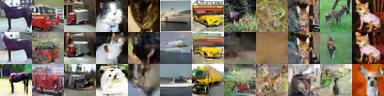}} \quad
    \subfigure{\includegraphics[width=0.30\textwidth]{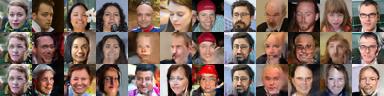}} \par
    \vspace{-3mm}
    \subfigure{\includegraphics[width=0.30\textwidth]{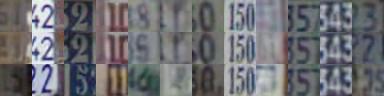}} \quad
    \subfigure{\includegraphics[width=0.30\textwidth]{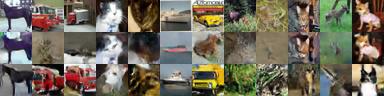}} \quad
    \subfigure{\includegraphics[width=0.30\textwidth]{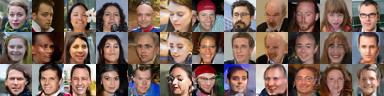}} \par
    \vspace{-3mm}
    \subfigure{\includegraphics[width=0.30\textwidth]{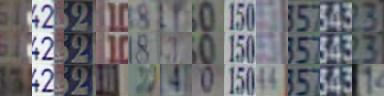}} \quad
    \subfigure{\includegraphics[width=0.30\textwidth]{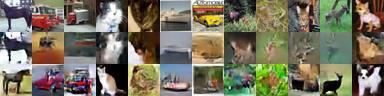}} \quad
    \subfigure{\includegraphics[width=0.30\textwidth]{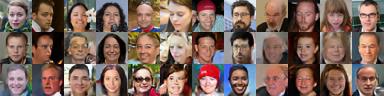}} \par
    \vspace{-3mm}
    \caption{Analysis of mode drop on SVHN (left), CIFAR10 (middle) and FFHQ (right) as a function of mini-batch size (3,200/800/200/50 from the top) and noisy latent space (bottom row). For each subfigure, the first row shows data samples, while the second and third rows show the closest generated samples in pixel space and latent space, respectively.}
    \label{fig:md}
    \vspace{-2mm}
\end{figure*}

In Fig.~\ref{fig:of}, larger batch sizes exhibit more overfitting (nearly indistinguishable samples), while smaller ones miss certain data points and produce more artifacts over the samples. Even though some of the artifacts are not easy to see, they influence the nearest neighbor in latent space. Similarly, Fig.~\ref{fig:md} shows mode drop behavior for various mini-batch sizes. The pattern closely resembles the overfitting results, where smaller batch sizes (more stochasticity) exhibit more mode drop, i.e.\ nearest neighbor search from data samples to generated ones does not return to similar samples. In both analyses, SVHN is affected less with respect to batch size change compared to CIFAR-10 and FFHQ, which are harder to model.

Quantitative results (pixel-wise loss) are shown in Table~\ref{quantitative_table}. Scores for overfitting and mode drop exhibit similar patterns and supports our qualitative observations: as stochasticity decreases overfitting occurs and mode drop diminishes. This shows that modeling all parts of the distribution (no mode drop) is not sacrificed for image generation quality (overfitting). However there is a trade-off between generalization and mode drop. Table~\ref{quantitative_table} also shows the scores for worst-case samples, which increase with lower percentages. As in the Figures, SVHN scores are less affected when batch size changes than the other datasets.

\begin{table}[h]
  \caption{Pixel-wise loss for overfitting and mode drop as a function of mini-batch size $m$ and noisy latent code (\textit{Noise}).}
  \vspace{1mm}
  \label{quantitative_table}
  \centering
  \resizebox{\columnwidth}{!}{
  \begin{tabular}{lcc|rrr|rrr}
    \toprule
    Dataset & $m$ & \textit{Noise} &\multicolumn{3}{c}{Overfitting} & \multicolumn{3}{c}{Mode Drop}\\
    \midrule
    & & & Avg & 10\% & 5\% & Avg & 10\% & 5\% \\
    \midrule
    SVHN & 3,200 & No & 1.55 & 2.82 & 3.58 & 1.55 & 2.50 & 2.80\\
    SVHN & 800 & No & 3.05 & 4.87 & 5.55 & 3.26 & 6.35 & 7.88\\
    SVHN & 200 & No & 7.43 & 11.59 & 12.42 & 7.84 & 12.66 & 13.99\\
    SVHN & 50 & No & 9.06 & 16.63 & 18.6 & 9.36 & 20.8 & 25.98\\
    SVHN & 3,200 & Yes & 6.09 & 10.71 & 11.85 & 7.15 & 18.07 & 22.76\\
    \midrule
    CIFAR10 & 3,200 & No & 4.90 & 7.40 & 8.59 & 4.88 & 6.90 & 7.39\\
    CIFAR10 & 800 & No & 8.04 & 12.96 & 14.51 & 8.43 & 13.73 & 15.6\\
    CIFAR10 & 200 & No & 24.47 & 37.01 & 38.87 & 24.71 & 40.44 & 43.36\\
    CIFAR10 & 50 & No & 28.31 & 42.42 & 44.35 & 29.49 & 47.37 & 50.89\\
    CIFAR10 & 3,200 & Yes & 34.39 & 49.57 & 51.41 & 32.16 & 48.25 & 51.31\\
    \midrule
    FFHQ & 3,200 & No & 6.39  &  9.71 & 11.81 & 6.57 & 12.55 & 17.41\\
    FFHQ & 800 & No & 10.7 & 16.2 & 18.96 & 10.72 & 17.64 & 21.74\\
    FFHQ & 200 & No & 27.81 & 38.78 & 40.78 & 28.38 & 39.57 & 41.59\\
    FFHQ & 50 & No & 32.84 & 43.79 & 45.19 & 32.57 & 44.37 & 46.27\\
    FFHQ & 3,200 & Yes & 35.56 & 45.83 & 47.53 & 33.9 & 44.51 & 46.04\\
    \bottomrule
  \end{tabular}
  }
\end{table}

\subsection{Noisy Latent Code}

We now show the effect of the second type of stochasticity, namely noisy latent code, to the generator. In our experiment one model receives noise to its input (as explained in Section \ref{experiments}), while the other does not. Otherwise they are the same, with a mini-batch size of 3,200. Both models are trained for the same amount of iterations, but the results are drastically different, especially for CIFAR10 and FFHQ (see Figs.~\ref{fig:of} and \ref{fig:md}, and Table~\ref{quantitative_table}). Overfitting analysis of noisy input (Fig.~\ref{fig:of} bottom row) can only show limited similarity between generated and training data (some ships and dogs in case of CIFAR10), while noise-free counterpart shows exact matching in each column. Similar behavior can be seen from the mode drop analysis (Fig.~\ref{fig:md}). SVHN results seem to be robust against the change in both analyses visually, but a significant difference can still be seen in Table~\ref{quantitative_table}. 


A comparison of the two types of stochasticity (Table~\ref{quantitative_table}) shows that noisy input influences overfitting and mode drop more than the noise coming from a batch-size of $50$ for CIFAR10 and FFHQ. Nevertheless, both types of stochasticity support our hypothesis. Furthermore, image generation does not seem to reach a satisfactory level of fidelity when noise is included, especially in CIFAR10. 

\section{Conclusions}
We have shown empirically that GANs can overfit and show little to no mode drop when stochasticity is removed from training. Moreover, we have shown the trade-off between generalization and mode drop; it can be adjusted by changing the amount of stochasticity. 
We believe our observations on the effects of stochasticity in GAN training can benefit GAN design criteria and future research on this topic. 


\noindent \textbf{Acknowledgement:} The computational work for this article was performed on resources of the National Supercomputing Centre, Singapore (https://www.nscc.sg).

\newpage
\bibliographystyle{IEEEtran}
\bibliography{strings,refs}

\newpage

\end{document}